\documentclass{article}

  \usepackage[preprint]{neurips_2025}

\usepackage{amsmath,amsfonts,amsthm,amssymb}

\usepackage[utf8]{inputenc} % allow utf-8 input
\usepackage[T1]{fontenc}    % use 8-bit T1 fonts
\usepackage{hyperref}       % hyperlinks
\usepackage{url}            % simple URL typesetting
\usepackage{booktabs}       % professional-quality tables
\usepackage{amsfonts}       % blackboard math symbols
\usepackage{nicefrac}       % compact symbols for 1/2, etc.
\usepackage{microtype}      % microtypography
\usepackage{xcolor}         % colors
\usepackage{graphicx}
\usepackage{caption}
\usepackage{subcaption}

\title{The View From Space: Navigating Instrumentation Differences with EOFMs}

\author{%
  Ryan P. Demilt\thanks{Cooresponding Author: rdemilt@sig-gis.com}, Nicholas LaHaye, Karis Tenneson\\
  Spatial Informatics Group, Pleasanton, CA 94566, USA
}

\begin{document}

\maketitle

\begin{abstract}
Earth Observation Foundation Models (EOFMs) have exploded in prevalence as tools for processing the massive volumes of remotely sensed and other earth observation data, and for delivering impact on the many essential earth monitoring tasks. An emerging trend posits using the outputs of pre-trained models as 'embeddings' which summarize high dimensional data to be used for generic tasks such as similarity search and content-specific queries. However, most EOFM models are trained only on single modalities of data and then applied or benchmarked by matching bands across different modalities. It is not clear from existing work what impact diverse sensor architectures have on the internal representations of the present suite of EOFMs. We show in this work that the representation space of EOFMs is highly sensitive to sensor architecture and that understanding this difference gives a vital perspective on the pitfalls of current EOFM design and signals for how to move forward as model developers, users, and a community guided by robust remote-sensing science.
\end{abstract}

\section{Introduction}

The remote sensing community has long driven impact through the skillful adoption and adaptation of machine learning techniques \cite{dl4rs_review}, particularly owing to the close relationship of many satellite image processing problems to developments in the computer vision domain. Recently, a paradigm shift has been proposed where in distinction to bespoke architectures and training procedures for each new task, models are proposed that are pre-trained on large collections, often entire mission archives worth, of data using self-supervised objectives. The data collections and objectives are carefully selected with the goal of learning discriminative, high-level but general features, producing 'Foundation Models' which have generic utility and can be used for a variety of downstream tasks. A diversity of models, largely based on variants of the ViT \cite{dosovitskiy2021imageworth16x16words} have been proposed for the task \cite{fuller2023cromaremotesensingrepresentations,wang2023ssl4eos12largescalemultimodalmultitemporal,prithvi,xiong2024neuralplasticityinspiredmultimodalfoundation,clayfoundationClayFoundation,jakubik2025terramindlargescalegenerativemultimodality}, each with bespoke data curation regimes, architectural modifications, and pretraining objectives; often with only one optical input stream used in pretraining.

In contrast to the relatively uniform space of traditional computer vision images, which share a consistent space of three spectral bands (RGB) and standard quantization (0-255), earth observation data is highly non-uniform. Sensors have diverse architectures and observation conditions. Compounding this problem are the complex processing stacks that are added for radiometric and geometric calibration, atmospheric correction, cloud-masking, and other considerations. Two sensors, observing the same location in roughly the same spectral range in the optical spectrum will likely register subtly different readings that may impact our broader image distributions. 

Large advances in evaluating EOFMs have been made recently: \cite{lacoste2023geobenchfoundationmodelsearth} \cite{pangaea}, each offer a diversity of tasks, application areas, geographic coverage, and modalities as well as thorough evaluation criteria. However, outside of raw performance numbers, the impact of sensor modality on the outputs pretrained models is under-discussed. It is not immediately obvious, for example, that a model pretrained on information collected by a Landsat-8 sensor should even be expected to carry significant symmetry over to samples from the Harmonized Landsat and Sentinel (HLS) \cite{hls} reflectance dataset, further into a subset of the same spectral bands collected by a Sentinel-2 satellite, or even beyond to more diverse architectures such as those that capture hyperspectral or extreme high-spatial-resolution imagery. These configurations are non-homogeneous and ofthen times require decades of their own development and engineering.

\section{Dataset}

To evaluate each model's response to diverse input streams we construct a dataset to dataset for comparison of the same points by many different views. To deepen evaluation it is also valuable to reference our findings to a supervised label which can be used to compare semantics between embedding spaces. To date, no benchmark dataset provides this capability as most datasets include only single modalities or only a single optical modality and SAR components. This is a non-trivial challenge as each modality will have a different acquisition period and therefore some work will need to be done to align these as best as possible.

To meet this challenge we design a collection procedure to generate images with paired samples from each modality. We randomly sample 600 points from the US State of Indiana and draw 224x224 pixel 2-month mosaic composite images at 30-meter spatial resolution for the three optical modalities (and four datasets) examined: HLS, Landsat-8, Landsat-9, Sentinel-2, and our single SAR modality, Sentinel-1. All optical data used is surface reflectance data, and SAR-GRD using the VV and VH polarizations of backscatter readings, in the Interferometric Wide Swath (IW) sensor mode. We utilize the QA pixels of each optical modality to mask images that are more than 20\% occluded by clouds. Finally we align the USDA Crop Data Layer as a reference to explore image semantics in our later analysis.

\section{Methodology}

In order to better evaluate the biases incorporated by the pretraining phase on the representation capabilities of our models we will restrict ourselves to an unsupervised analysis utilizing encoders with their weights frozen. Many interesting questions can be further explored in the realm of understanding the impact fine-tuning has on the weights and representations of the models used and on the downstream performance implications in specific cases. We propose these as future work.

To perform our evaluations we will utilize the Prithvi \cite{prithvi} and DOFA \cite{xiong2024neuralplasticityinspiredmultimodalfoundation} foundation models, as representative examples of the broad category of EOFMs with specific characteristics that we will highlight for their utility to our analysis. 

The Prithvi model is useful to us because it takes a six spectral band input and was pretrained solely using HLS data. The six input bands from each data stream can be accepted by matching the spectral bands by wavelength categories and discarding the additional bands. This can be seen as archetypal of many EOFM models.

The DOFA model is notable for its use of a dynamic network embedding layer, which utilizes the central wavelengths of the inputs to generate the weights for a convolution-based input layer. The model then processes the image. This allows the model to accept inputs with any band configuration and generate a valid output. However, the model characterizes inputs solely by their central wavelength. The model has also been trained with multiple modalities of data from aerial RGB to satellite data. The flexibility and multimodal training combine to offer a useful counterpoint to the standard for many models.

\subsection{Embedding Space Visualization}

The high dimensional nature of the models' latent spaces, 768 for the selected models, makes sample variability difficult to evaluate and impossible to visualize in their native form. Visualization and dimensionality reduction techniques such as TSNE \cite{JMLR:v9:vandermaaten08a}, offers a perspective on the internal structures of these spaces, and the major axes on which data varies from each other. This makes the technique well suited to our comparative analysis. Ideally, there would not be identifiable clusters based on modality, in either the original heigh-dimensional space or in the low dimensionality reduction, as this would imply that in our high dimensional space data is greatly or principally distributed by factors other than data source. This would be shown best by our embedding spaces being segmentable by semantic features of the image collection such as land cover, crop types, or other differentiating phenomena. 

\subsection{Local Neighborhood Analysis}

The second method for comparing our embedding spaces as we vary input modality is to consider the local neighborhoods of each point based on distance in the embedding space. This is based on the intuitive notion that similarity between data points is a perspective on the overall configuration of the space \cite{Duderstadt2023ComparingFM,tavares2024measuringsimilarityembeddingspaces} . We can evaluate the similarity of either image patch embeddings or cls tokens. Patch embeddings will evaluate the differentiation in fine grained detail and the cls token should give a summary view of our images. We consider this to be a valuable perspective for two reasons. First, EOFM embeddings of satellite data have been proposed for few-shot and similarity search contexts where the ease of separability and evaluation for embeddings of data has important implications \cite{Blumenstiel2024MultiSpectralRS}. Second, we expect that the subtle distributional differences between our modalities to cause some noise in absolute values of the embedding spaces and at high dimensions these distances are very large in absolute magnitude, by considering neighborhoods we can establish a baseline of what is indicated as similar by a pretrained model and the sensitivity to the sensor architecture.

\section{Results}

\begin{figure}
    \centering
    \captionsetup{width=.4\linewidth}
    \begin{minipage}{.5\linewidth}
        \centering
        \includegraphics[width=\linewidth]{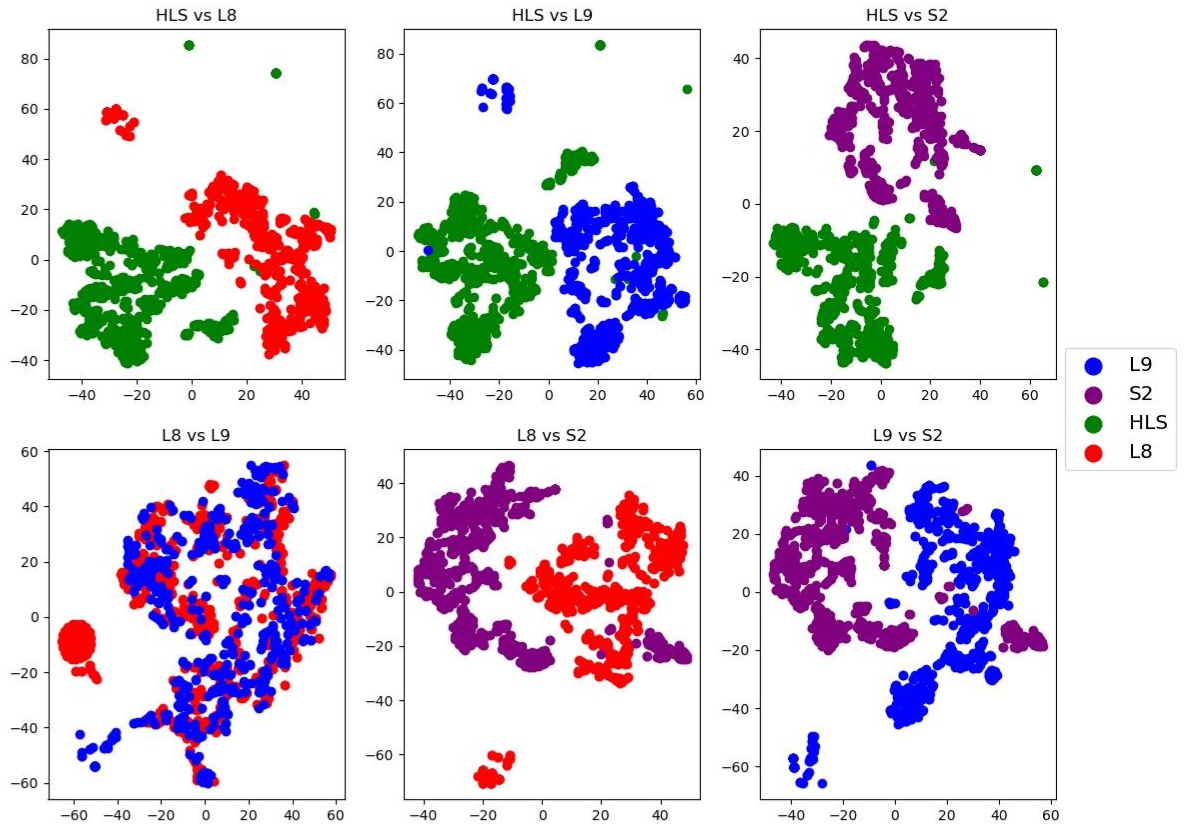}
        \captionof{figure}{Prithvi TSNE Plots optical modality embeddings}
        \label{fig:prithvi_tsne}
    \end{minipage}%
    \begin{minipage}{.5\linewidth}
        \centering
        \includegraphics[width=\linewidth]{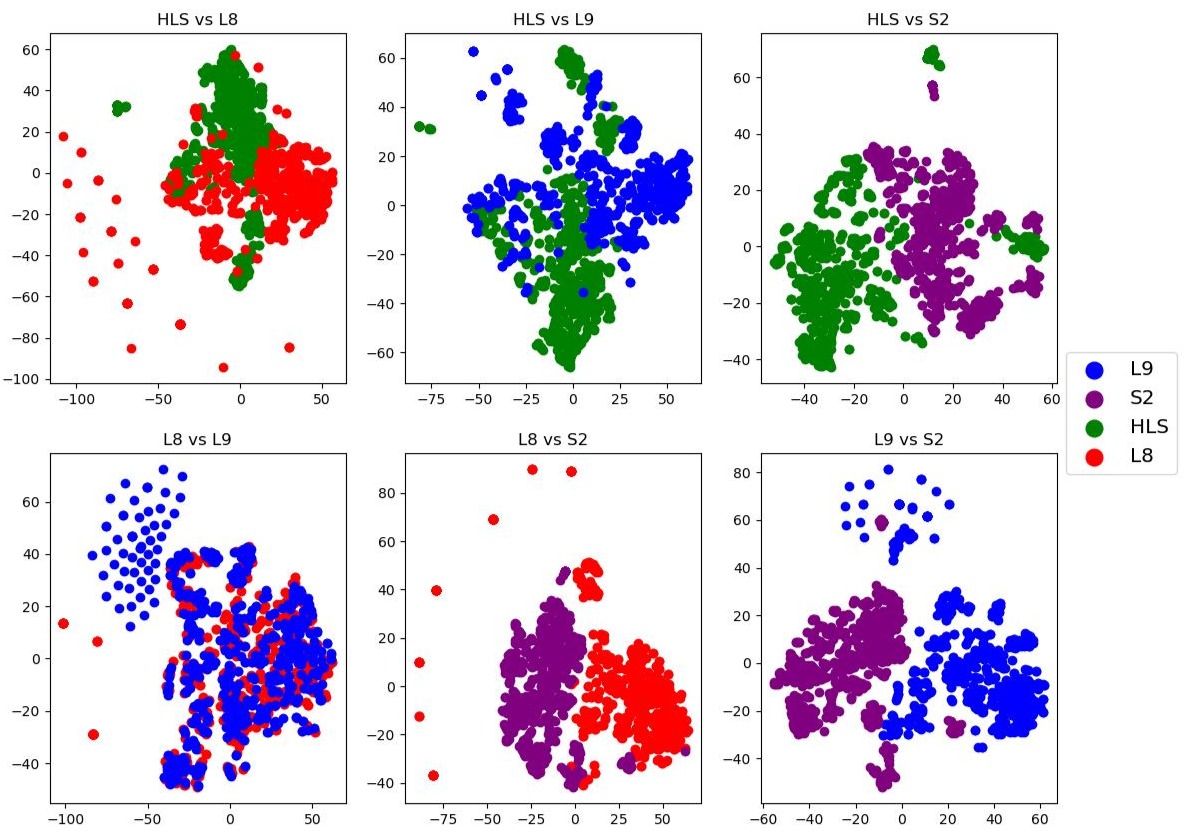}
        \captionof{figure}{DOFA TSNE Plots optical modality embeddings}
        \label{fig:dofa_tsne}
    \end{minipage}
    \vspace{-2mm}
\end{figure}

To use TSNE to visualize the embedding space in a low dimension, we plot the TSNE transformation of each point and color its source modality. We find in Figures \ref{fig:prithvi_tsne} \& \ref{fig:dofa_tsne} that between most pairs of modalities there are clear clusters of points based on their source modality. The clear separations show that in many cases the minor differences in the distributions of instrument readings are resulting in large variances. It is unsurprising that many of the strongest performing pairings for mixing data of each modality involve the HLS data which is a uniform synthesis of the observations from the other optical data streams and the Landsat pairing which carry nearly identical sensors. The DOFA architecture shows a relatively greater resilience to modality and source change in some cases, compared to Prithvi, but still shows distinct separations in many pairwise cases. 

% \begin{figure}
%     \centering
%     \includegraphics[width=0.5\linewidth]{figures/prithvi_tsne_plots.jpg}
%     \caption{Pairwise TSNE Plots for optical modalities using Prithvi model embedding}
%     \label{fig:tsne}
% \end{figure}

\begin{table}

\begin{tabular}{|c|c|c|c|c|c|} \hline
\small{Modalities} & \small{Prithvi \# Matches} & \small{Prithvi \% Matches} & \small{Dofa \# Matches} & \small{Dofa \% Matches} \\ \hline
\small{Landsat 8} \& \small{Landsat 9} & \small{1 / 2 / 4}  & \small{17.7 / 17.8 / 21.1} & \small{1 / 2 / 5} & \small{20.60 / 22.1 / 23.9}\\ \hline
\small{Landsat 8} \& \small{HLS} & \small{1 / 2 / 5} & \small{22.3 / 24.0 / 26.8} & \small{1 / 2 / 5} & \small{23.6 / 24.3 / 27.2}\\ \hline
\small{Landsat 8} \& \small{Sentinel-2} & \small{1 / 2 / 4} & \small{13.6 / 15.8 / 18.9} & \small{1 / 2 / 3} & \small{17.4 / 18.6 / 20.4} \\ \hline
\small{Landsat 9} \& \small{HLS} & \small{1 / 3 / 7} & \small{28.5 / 30. 5/ 33.1} & \small{1 / 3 / 6}& \small{29.7 / 28.2 / 29.3}\\ \hline
\small{Landsat 9} \& \small{Sentinel-2} & \small{1 / 2 / 4} & \small{16.8 / 18.8 / 22.3} & \small{1 / 2 / 5}& \small{18.1 / 20.3 / 23.1}\\ \hline
\small{HLS} \& \small{Sentinel-2} & \small{1 / 2 / 5} & \small{19.0 / 23.1 / 26.7} & \small{1 / 2 / 5}& \small{21.0 / 21.9 / 25.0}\\ \hline
\end{tabular}
\caption{Average number of matches (rounded) and average \% of local neighborhood agreement between embedded modalities. Each row shows the matches for 5, 10, and 20 nearest neighbors respectively, separated by the "/".}
\label{tab:cls_sim}
\vspace{-4mm}
\end{table}

Transitioning to our second evaluation perspective we examine the local neighborhoods of cls tokens for both models. In Table \ref{tab:cls_sim} we show another pairwise view of our modalities. Each row lists the rounded number of common neighbors in a 5, 10, and 20-Nearest Neighbors list betweeen two modalities and the average percentage of matching neighbors across the dataset. We find that the percentage of our matching neighbors in local neighborhoods of our cls tokens is only one case greater than 30\% on average, meaning that in a context such as similarity search, input modality becomes critically to downstream performance and the determinant of the majority of similar candidates.  

Next we examine the neighborhood of the patch embeddings rather than cls tokens. We take the patch embeddings of every patch in our dataset and compare to find the 10-Nearest Neighbors by cosine similarity. As a proxy for semantic similarity at a patch level we calculate the percentage of these neighbors which share a most common class with the original point, using the USDA Crop Data Layer as labels. The average of these over all patches is presented in Table \ref{tab:neighbors}. The results show as much as a 4\% difference in shared most common class depending on the modality chosen. This provides further evidence that modality change causes not only neighborhood structure to shift but also creates changes in semantic similarities between local image patches. In a search system this could result in vastly different recall results for a query, or different classification in the few/zero-shot settings. Table \ref{tab:modality_prediction} shows a complementary perspective wherein two simple classifiers, a RandomForest with 500 estimators and a 5-Nearest Neighbor Classifier, are tasked with predicting the input modality for a patch. We find that with 25\% of our patches held out as a test set a simple Random Forest classifier can still get a shocking 90.7\% and 88.7\% accuracy for Prithvi and DOFA embeddings respectively in predicting which modality a patch embedding originated from, implying it is relatively simple to partition the embedding space of both models into where samples from each modality are placed.

% \begin{table}[]
%     \centering
%     \begin{tabular}{|c|c|c|c|}
%     \hline
%         Modality & Random & Prithvi & DOFA \\ \hline
%         Landsat 8 & 24.1 & 47.3 & 49.0  \\ \hline
%         Landsat 9 & 24.1 & 56.1 & 68.5\\ \hline
%         HLS & 24.1 & 59.5 & 72.1\\ \hline
%         Sentinel-2 & 24.1 & 55.6 & 70.5\\ \hline
%         Difference & - & 12.2 & 23.1 \\ \hline
%     \end{tabular}
%     \caption{Average Percentage of 10-Nearest Neighborhood with the same mode Crop class for embedded patches in crop data layer labels}
%     \label{tab:neighbors}
% \end{table}

\begin{table}
\centering
\captionsetup{width=.4\linewidth}
\begin{minipage}{.5\linewidth}
    \centering
    \begin{tabular}{|c|c|c|c|}
    \hline
        Modality & Random & Prithvi & DOFA \\ \hline
        Landsat 8 & 24.1 & 57.0 & 69.3  \\ \hline
        Landsat 9 & 24.1 & 57.6 & 70.8\\ \hline
        HLS & 24.1 & 59.5 & 71.9\\ \hline
        Sentinel-2 & 24.1 & 55.5 & 69.3\\ \hline
        Difference & - & 4.0 & 2.6 \\ \hline
    \end{tabular}
    \caption{Average Percentage of 10-Nearest Neighborhood with the same mode Crop class for embedded patches in crop data layer labels}
    \label{tab:neighbors}
\end{minipage}%
\begin{minipage}{.5\linewidth}
    \centering
    \begin{tabular}{|c|c|c|}
    \hline
        EOFM & RF Accuracy & KNN Accuracy \\ \hline
        Prithvi & 90.7\%  & 85.6\%\\ \hline
        DOFA & 88.7\% & 91.1\%\\ \hline
    \end{tabular}
    \caption{Prediction Accuracy for modality source of patch embed using Random Forest \& 5-NN Classifier}
    \label{tab:modality_prediction}
\end{minipage}
\vspace{-4mm}
\end{table}

\section{Discussion}

We find that input modality has a large impact on the outputs of EOFMs. Along these findings we recommend that users of EOFM models pay careful mind to their data collection regime and its alignment with the pretraining data of the chosen model. Combined with the findings in \cite{pangaea}, we propose that it is likely matching spectral bands by wavelength and training on single modalities are insufficient for understanding the complexities of spectral capture technology that extend to subtler underlying distribution shifts than simply matching wavelengths against each other.

Our findings motivate consideration of models based on the sources of their pretraining data and on consideration of diverse sensor modalities for the complexity that clearly underlies them. Additionally we hope this motivates the creation of further highly multimodal and multi-input stream datasets such as our example here to test the cross-modality capabilities of new models.

\section{Data Availability Statement}

We utilize the model versions available through PANGAEA \cite{pangaea} to generate embeddings of our multi-modal data. The dataset is available online along with the code used to generate all figures which can be accessed at the following \href{https://anonymous.4open.science/r/Navigating-Instrumentation-Differences-with-EOFMs-7B0D}{GitHub repository}. 

\bibliography{references}

\end{document}